\documentclass[10pt,twocolumn]{ICCAS2024}
 
\usepackage{diagbox}
\usepackage{amsmath} 
\usepackage{amssymb}  
\usepackage{lmodern}
\usepackage{array}
\usepackage{moresize}
\usepackage{tabularx}

\usepackage{algorithmic, algorithm} 

\usepackage[inline]{enumitem}

\usepackage{cite}
\usepackage{float}
\usepackage{placeins}
\usepackage{hyperref}
\usepackage{graphicx}
\usepackage{booktabs}
\graphicspath{{./figures/}}

\begin{document}

\title{STAR: Swarm Technology for Aerial Robotics Research}

\author{
    Jimmy Chiun${}^{1*}$, Yan Rui Tan${}^{1}$, Yuhong Cao${}^{1}$, John Tan${}^{1}$, Guillaume Sartoretti${}^{1}$ 
}

\affils{ ${}^{1}$College of Design and Engineering, National University of Singapore, Singapore \\
\{jimmy.chiun, yr.tan, johntgz\}@u.nus.edu, \{caoyuhong, mpegas\}@nus.edu.sg \\}

\thanks{\small${}^{*}$ Corresponding author}

\abstract{
    In recent years, the field of aerial robotics has witnessed significant progress, finding applications in diverse domains, including post-disaster search and rescue operations. Despite these strides, the prohibitive acquisition costs associated with deploying physical multi-UAV systems have posed challenges, impeding their widespread utilization in research endeavors. To overcome these challenges, we present STAR (Swarm Technology for Aerial Robotics Research), a framework developed explicitly to improve the accessibility of aerial swarm research experiments. Our framework introduces a swarm architecture based on the Crazyflie, a low-cost, open-source, palm-sized aerial platform, well suited for experimental swarm algorithms. To augment cost-effectiveness and mitigate the limitations of employing low-cost robots in experiments, we propose a landmark-based localization module leveraging fiducial markers. This module, also serving as a target detection module, enhances the adaptability and versatility of the framework. Additionally, collision and obstacle avoidance are implemented through velocity obstacles. The presented work strives to bridge the gap between theoretical advances and tangible implementations, thus fostering progress in the field.
}

\keywords{
    Aerial Robotics, Micro UAV, Landmark Localization, Robotic Application
}

\maketitle


\section{Introduction}
\label{sec:introduction}

In recent times, autonomous robots have experienced a steady increase in their penetration in various domains. In industrial settings, their utilisation is increasingly becoming more prevalent in tasks such as manufacturing and logistics\cite{queralta2020collaborative}. However, the adoption of autonomous robots in practical scenarios is still laden with challenges such as high acquisition costs, deployment in unknown indoor environments, and adapting to the high cost and technological difficulties of deploying large fleets of aerial robots often hinder experimental validation. That is, most work still focuses on simulations \cite{li2018sensor, zhen_intelligent_2020}, and physical experimental implementations often experience issues such as inaccurate onboard localization \cite{McGuire2019}. Also, most research quadcopters are large and need large space to operate, therefore small quadcopters became an attractive alternative to operate large swarms. Therefore, based on the work of \cite{crazyswarm2}, we propose STAR, a framework for indoor physical swarm experiments that aims to address these limitations. This paper provides a summary of our framework and discusses its significance for the aerial swarm research community.

The main contributions of this paper are as follows:

\begin{figure}[t]
    \centering   \includegraphics[width=0.46\textwidth]{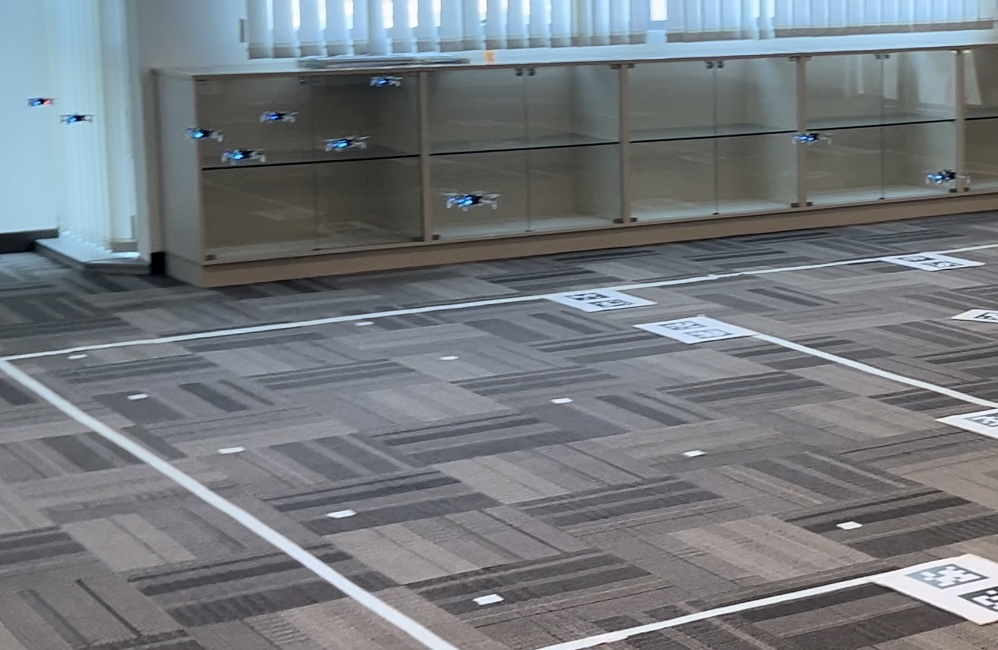}
    \caption{
     \textbf{Snapshot of aerial swarm using STAR}}
    \label{fig: STAR}
    \vspace{-0.65cm}
\end{figure}

\begin{itemize}
    \item A robust swarm management framework utilizing ROS2 for managing physical aerial swarms with integrated task manager and collision avoidance.
    \item We implemented a versatile landmark-based localisation module that is cost effective and allow the swarm system to be independent of expensive motion capture system. This module also provide target tracking capabilities.
    \item We developed a simulation environment that accurately simulates both the environment and landmark markers tracking. Thus reducing the gap between simulated and real-world performance through an improved and realistic custom camera model.
    \item We provide validation results for the proposed landmark-based localisation approach, showing that it effectively limits UAV localisation errors.
    \item Our swarm framework is made available as open source to the robotics research community\footnote{Source code available at: https://github.com/STAR-swarm/STAR}.
\end{itemize}

\section{Related works}
\label{sec:related_works}
Aerial swarms have been experimented with for various purposes such as formation flight, trajectory control, and numerous works that has been proposed for operating them. \cite{aerostack2_2023} described a software framework for aerial robotics that aims to streamline the development of multi-robot systems. By leveraging ROS 2 middleware and a modular software architecture, Aerostack2 offers platform independence, versatility, and multi-robot orientation, making it suitable for a wide range of UAV applications. Although the framework covers a wide range of robot capabilities, there may be scenarios in which additional customization or integration with specific hardware components is necessary. 

\cite{crazychoir_2023} introduces a ROS 2 toolbox tailored for swarms of Crazyflie nano-quadrotors, enabling realistic simulations and experiments. It provides a modular framework for fast prototyping of decision-making and control schemes, handling each Crazyflie with independent ROS 2 processes to reduce single points of failure. By integrating with Webots and firmware bindings, CRAZYCHOIR enables distributed optimization, cooperative decision-making, and communication among nano-quadrotors. 

Another notable work is Crazyswarm \cite{crazyswarm}, which features a robust and synchronized control architecture for large indoor quadcopter swarms. Built on the Crazyflie platform, it leverages onboard computation to ensure reliability against communication issues while requiring minimal radio bandwidth. The system demonstrates good scalability in terms of latency and tracking performance as the swarm size increases. However, these works face limitations, such as dependence on a motion-capture system, which serves as a single point of failure, and the high acquisition cost. This presents a fundamental challenge for operating UAVs experimentally in indoor environments. Indoor localization methods include ultra-wideband (UWB)~\cite{chu2022swarm} or external anchor systems like \cite{taffanel2021lighthouse}. The high cost of these systems can hinder experimental validation of aerial swarm algorithms. Additionally, these systems depend on extra radios and sensors for localization, which reduce the UAV's endurance without adding any mission capability \cite{McGuire2019}. All these frameworks rely on motion capture systems, restricting the design of cluttered environments as they might obstruct the cameras.

In addition to the challenge of localization, collision avoidance (CA) emerges as a significant hurdle. The CA problem is especially challenging for nano-sized quadrotors due to limited payload, sensing, and computing power.\cite{Kang2019} described a hybrid deep reinforcement learning (DRL) approach that combined real-world and simulation data to avoid static obstacles with a monocular camera mounted on a Crazyflie. Predictive control-based model approaches were studied in\cite{Jin2022} with indoor experiments using Crazyflies.

\section{Framework Overview}
\label{sec:framework_overview}

\begin{figure}
    \includegraphics[width=\columnwidth]{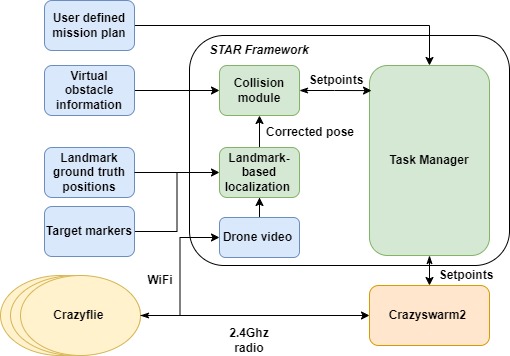}
    \caption{\textbf{STAR Framework Overview}}
    \label{fig:task_manager}
    \vspace{-0.7cm}
\end{figure}

We consider the problem of coordinating a team of aerial robots with the ability to conduct a range of mission, which includes but not limited to target search and path planning. We are interested in the development of an integrated swarm framework that spans the entire mission spectrum, commencing from the take-off phase and culminating in the landing stage. Fig. ~\ref{fig:task_manager} describes our STAR framework. 

In the framework, we leverage the Crazyswarm2 communication protocol \cite{crazyswarm2} to handle communications between the ground computer and the swarm. Building upon this communication layer, we developed an application layer that allows user to formulate mission plan in an unified manner. The application layer also implements landmark-based localization as well as offers target detection, which allows researchers to craft mission such as target search scenarios. To minimize sim2real gap, we also provide a lightweight simulator, which visualize the obstacles, landmark marker placement, as well as the camera field of vision. This enabled researchers to conduct a sanity check of their code and framework configuration before conducting physical experiments, thereby reducing the errors encountered during physical experiments.

\subsection{Vehicle}
Our framework utilizes the readily available and low-cost Bitcraze Crazyflie 2.1, equipped with an STM32F405 MCU for inertial state estimation and motor control onboard. Enhancements include the Bitcraze Flow deck v2, featuring a TOF sensor for height estimation and an optical flow sensor for motion computation, enabling 3D pose estimation. We also attached the AI Deck 1.1, which adds a GAP8 SoC, a monochrome camera, and an ESP32 WiFi chip for streaming images to the Ground Control Station. The Crazyflie, compact and ideal for indoor flights, excels in tight formations and large numbers, and its low inertia minimizes injury risk during crashes.

\subsection{Communications}
Each robot is equipped with the ability to transmit telemetry data to the ground control station (GCS) via a 2.4GHz radio. For communication with the UAV while it is on the ground, we employ the request-response technique from \cite{crazyswarm2} to set parameters like controller gains. During flight, the framework uses the broadcast communication method from \cite{crazyswarm2} to send its position and various mission commands, including take-off, landing, and the next setpoint. Additionally, we use WiFi-based communication to transmit visual observations from the onboard camera to the GCS. 

\section{Task Manager}
\label{sec:swarm_manager} 

The Task Manager is the central coordinator that unifies algorithmic elements and guides task execution within the system. Its functions encompass:

\begin{itemize}
  \item Overseeing swarm operations (agent initiation, status updates, etc.)
  \item Implementing fundamental swarm directives (take-off, movement, landing, etc.) and predefined/scripted scenarios
  \item Fusing swarm reasoning and algorithm integration (task distribution, collision and static obstacle evasion, etc.)
  \item Execution of instructions from external modules, such as a search planner for target search mission.
\end{itemize}

The key elements of the Swarm Manager and their integration are depicted in Fig.~\ref{fig:task_manager}. In our framework, the developer can create a mission plan outlining the tasks that the swarms need to accomplish during a specific mission in a single yaml file. These tasks could be specified for a single UAV or for the whole swarm. These mission plans are defined as actions that the robot will perform autonomously, such as following the commands of a search planner. 

\section{Collision Avoidance}
In our framework, the UAV receives set point that it needs to follow, and the collision avoidance module handles path planning. This involves generating intermediate waypoints to reach the given set points. Consequently, a lower-level velocity controller is responsible for navigating to these intermediate waypoints, taking into account both static obstacle avoidance and inter-agent collision prevention. In scenarios without obstacles, the velocity controller directs the agent using a velocity vector aimed directly at the intermediate waypoint. To ensure safety, we implemented Optimal Reciprocal Collision Avoidance (ORCA) for static obstacle avoidance and inter-agent collision prevention\cite{van2011reciprocal}. Due to the sensing limitations of UAVs, this framework primarily considers virtual obstacles. However, inter-UAV collision avoidance is calculated in real-time using actual UAVs' position data.

\begin{figure}[t]
    \centering   \includegraphics[width=0.48\textwidth]{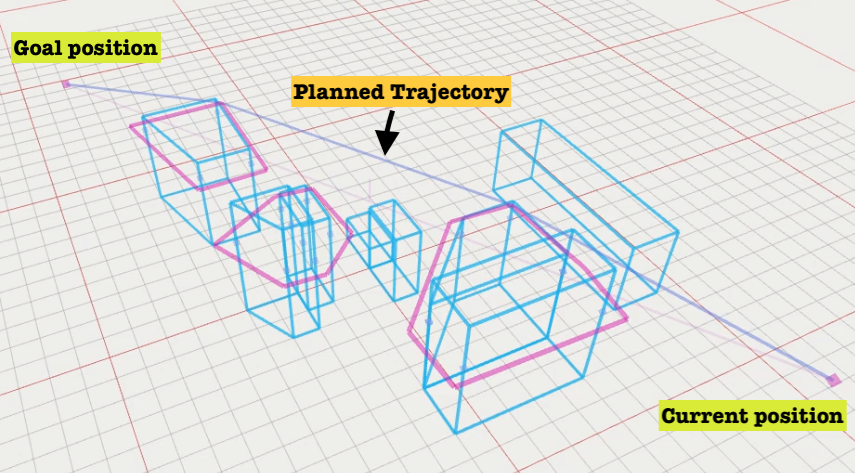}
    \caption{ \textbf{Illustration of collision avoidance.}
    2D polygonal surfaces (pink edges) can be derived from 3D obstacle maps, with virtual agents at key points along these edges. The ORCA algorithm can then be used for static obstacle avoidance. }
    \label{fig: ORCA3D}
    \vspace{-0.55cm}
\end{figure}

ORCA determines an alternative safe velocity close to the intended velocity by considering Velocity Obstacles (VOs)  (i.e., the vector pointing directly towards the intermediate waypoint). VOs delineate the potential collision range between two agents, denoted as $VO_{A|B}^{\tau}$, within a specified duration, $\tau$. If the current relative velocity, $v_{A|B}$, lies within $VO_{A|B}^{\tau}$, ORCA identifies a viable velocity, $orca_{A|B}^{\tau}$, closest to the intended velocity while avoiding collisions. In multi-agent scenarios, a sector of feasible velocities, $ORCA_{A|B}^{\tau}$, tangential to $VO_{A|B}^{\tau}$ at $orca_{A|B}^{\tau}$, is defined. This transforms collision avoidance into a linear programming problem (see Equations \ref{eqn:constraints} and \ref{eqn:cost}), aiming to select an optimal feasible velocity, $v_A^{new}$, closest to the intended velocity, $v_A^{des}$, within the intersection of all ORCA-defined sectors, $ORCA_{A}^{\tau}$ .

\begin{equation} \label{eqn:constraints}
    ORCA_{A}^{\tau} = \bigcap_{B \neq A} {ORCA_{A|B}^{\tau}}
\end{equation}

\begin{equation} \label{eqn:cost}
    v_A^{new} = \arg \min_{v \in {ORCA_A^{\tau}}} \lVert v - v_A^{des} \rVert
\end{equation}

ORCA can also be applied for static obstacle avoidance (see Fig. \ref{fig: ORCA3D}). Nonetheless, the complexity arises from the need to effectively define VOs in a manner that avoids excessive conservative avoidance, especially concerning large obstacles. To address this, a pragmatic approach is to model obstacles as polygons outlined by line segments. By identifying interpolated points along these edges that are significant for collision prevention. These interpolated points can then be treated as fixed virtual agents, augmenting ORCA for inter-agent collision avoidance. Note that, we use visibility graphs to calculate these points efficiently.

\section{Landmark-based localisation}

In our proposed framework, we use AprilTags as landmark markers, a visual fiducial system similar to QR codes, and address the localization problem as landmark-based SLAM with the GTSAM toolbox \cite{gtsam}. AprilTags can be easily printed and affixed, offering robust and affordable localization aids. Due to the inherent attributes of an AprilTag, its ID and predefined 6 DOF pose relative to the camera, the UAV can accurately determine its own pose and orientation in relation to the AprilTag.

\begin{figure}
    \centering
    \includegraphics[width=0.3\textwidth]{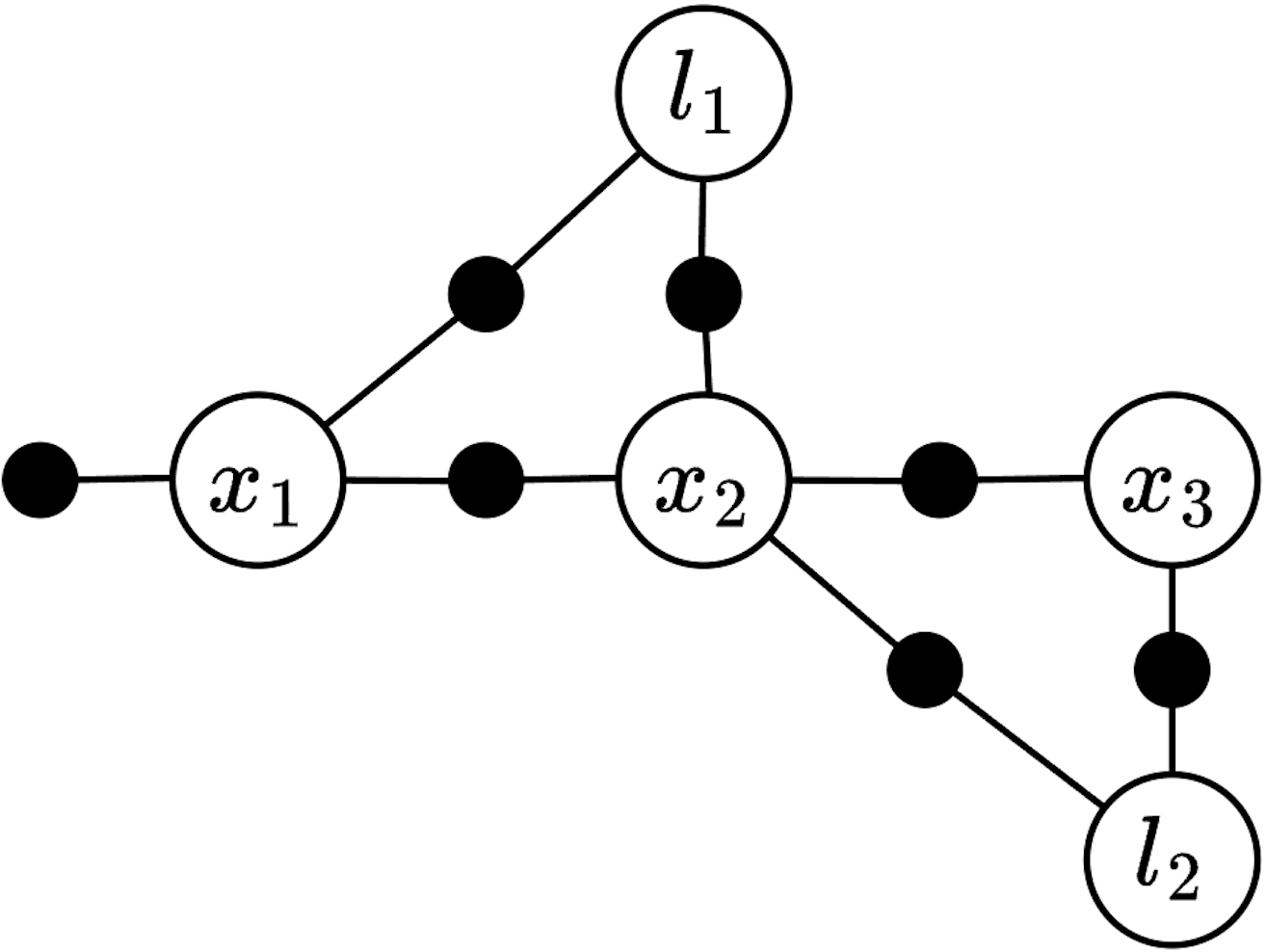}
    \label{fig:landmark_factor_graph}
    \caption{\textbf{Landmark-based SLAM factor graph}}
    \vspace{-0.55cm}
\end{figure}

Given that the global coordinates of the AprilTag landmarks are known a priori, we encode the Apriltag poses as an observed landmark $l_i$ into our factor graph (see Fig.\ref{fig:landmark_factor_graph}). The EKF estimated pose computed from fusing the IMU, optical flow and rangefinder measurements, are represented by $x_i$ and the solid black circles are factors which represent the constraints defined by the change in pose between each time step. 
The landmark SLAM problem then minimizes the non-linear error between all the nodes in the factor graph using algorithm \ref{alg:nonlin_opt_factor_graph} \cite{factor_graphs_for_robot_perception}. 

\vspace{0.3cm}
\begin{algorithm}
\begin{algorithmic}[1]
    \STATE  Given initial pose estimate $\tau^0$
    \WHILE {Error $J(\tau) = \sum_k \frac{1}{2} {\lVert log(\tilde{T}_{ij}^{-1} {T_i}^{-1} T_j ) \rVert}^2$ not converged}
        \STATE \hspace{\algorithmicindent}  Linearize the factors $\frac{1}{2} {\lVert log(\tilde{T}_{ij}^{-1} {T_i}^{-1} T_j ) \rVert}^2 \approx \frac{1}{2} {\rVert A_i \xi_i + A_j \xi_j - b \rVert}^2$
        \STATE \hspace{\algorithmicindent}  Solve the least squares problem ${\xi_i}^* = {argmin}_{\xi} \sum_k \frac{1}{2} {\lVert A_i \xi_i + A_j \xi_j - b \rVert}^2  $
        \STATE \hspace{\algorithmicindent}  Update pose variable $X^{t+1}_{i} \leftarrow X^t_j \Delta (\xi_i)$
    \ENDWHILE
\end{algorithmic}
\caption{Nonlinear optimization for factor graph: $\tau^0$: Initial set of estimated poses, $\tilde{T}_{ij}:$ Pose constraints between poses $T_i$ and $T_j$, $\xi$: incremental local coordinates, $A_i$: Jacobian matrix, $b$: bias term, $X_i$: pose variable}\label{alg:nonlin_opt_factor_graph}
\end{algorithm}
\vspace{0.2cm}

Consequently, this empowers the UAV with the capability to re-localize itself by effectively situating its position within the broader environmental context. To enhance localization accuracy, we employ a dual marker setup at each landmark point. This configuration provides the UAV a pair of reference points to rectify its positioning.  
Beyond acting as a landmark, AprilTags can also represent targets such as survivors or checkpoints which act as mission objectives.

\section{Simulation tool}
\begin{figure}
    \includegraphics[width=\columnwidth]{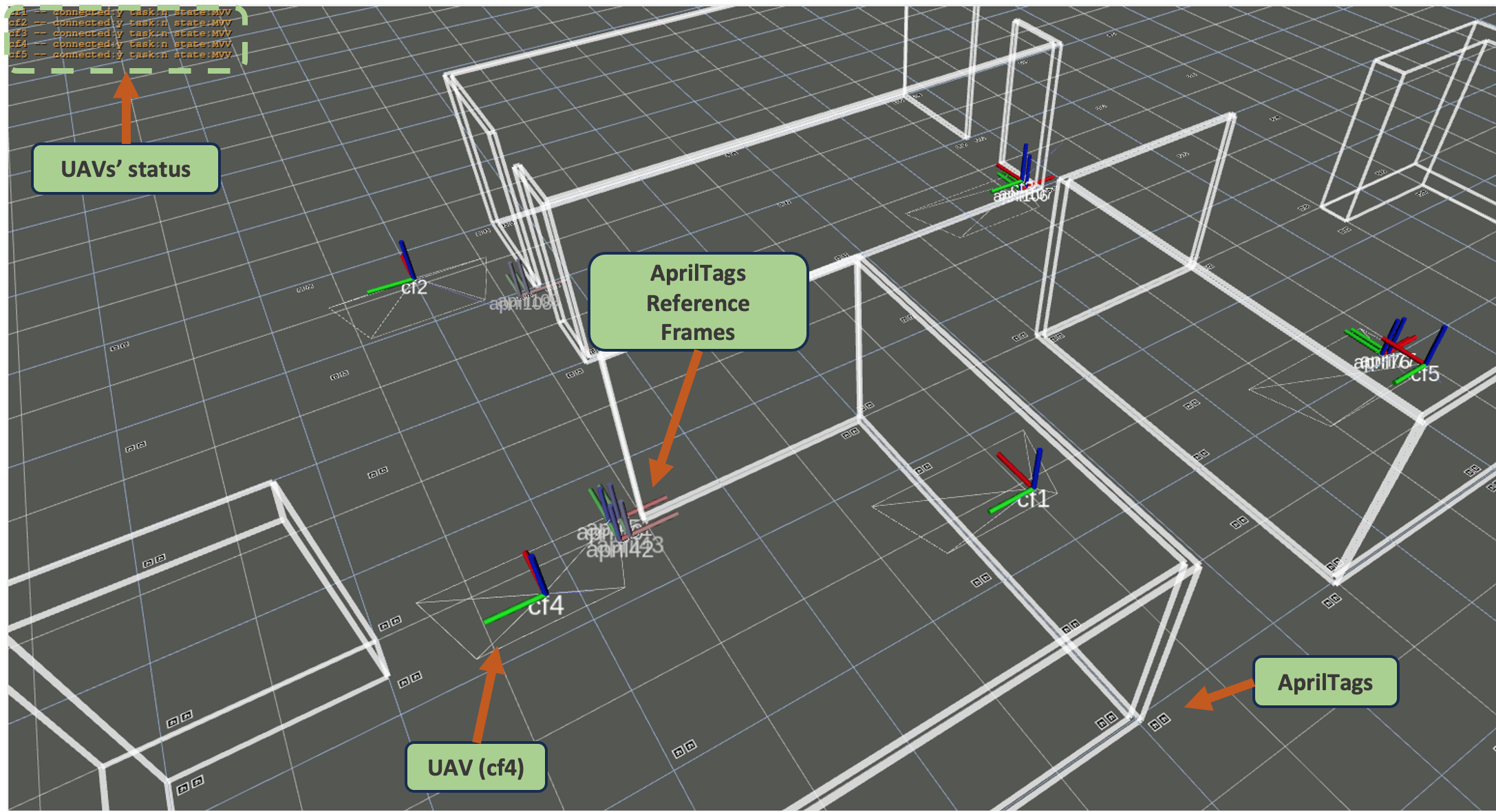}
    \caption{\textbf{Illustration of simulation environment:} Each UAV's orientation is depicted as a coordinate axis (with red representing the x-axis, green for y, and blue for z). The camera model is illustrated by white trapeziums.}
    \label{fig:safmc_arena}
    \vspace{-0.65cm}
\end{figure}

A lightweight simulator is also developed to provide a simulation environment for researchers to validate their algorithms and framework configurations before hardware deployment. The simulation is visualized through RVIZ2, while the computation is achieved by leveraging the Crazyflie's firmware python bindings and numerical integration \cite{crazychoir_2023}. We also visualised the obstacles as wire-frame objects as well as the landmark markers. This allows the developer to validate the the configuration file used is the intended experiment setup, i.e., position of landmark markers are coded in the framework as desired in the physical setup (see Fig. ~\ref{fig:safmc_arena}). This could be easily adjust based on the developer test environment by adjusting the config yaml file of the software.
\section{Experiment} 
\label{sec:experiments}
We conducted experiments to measure the various performance metrics of our framework, as well as to validate the landmark localization module.

\subsection{Experiment setup} 
\label{sec:experimental_results}

\begin{figure}
    \centering
    \includegraphics[width=0.46\textwidth]{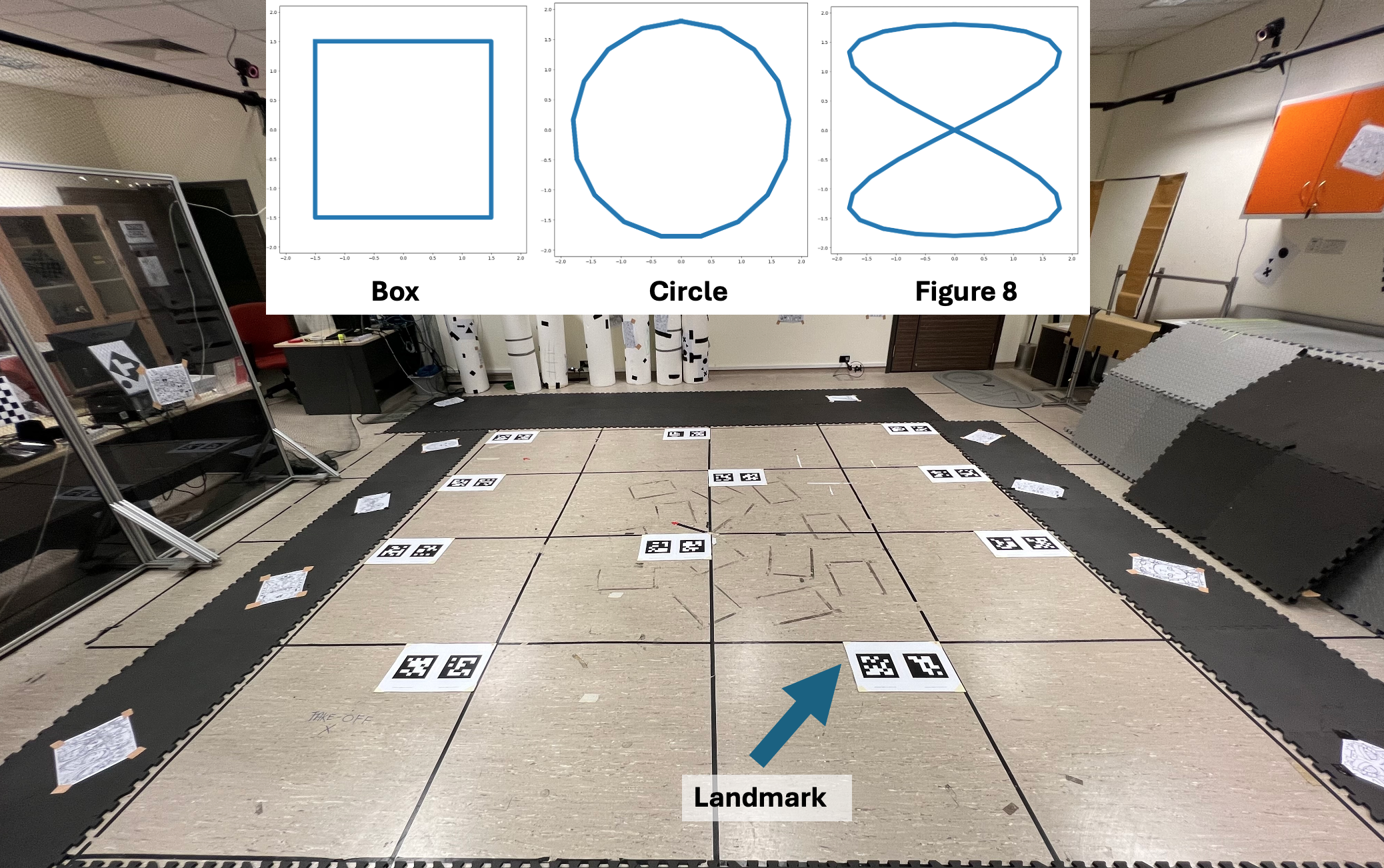}
    \caption{
    \textbf{Experiment setup:} Each box is 1m by 1m and flying area is 4m by 4m. Top: Types of test trajectories)}
    \label{fig:test_venue}
    \vspace{-0.5cm}
\end{figure}

We conducted an experimental validation of the real-world performance of the landmark localisation module by executing flights along three predetermined flight paths:  \textit{Box}, \textit{Circle} and \textit{Figure 8}  patterns (see Fig. \ref{fig:test_venue}). For each of the trajectory, \textit{three} laps were clocked. The actual travelling distance for each trajectory is 28.41m, 37.16m and 50.32m for the box, circle and Figure 8 respectively. To measure the mean square error of the position estimate computed by the landmark localization module, we also record the ground-truth position of the UAV using position estimates from the VICON system. During the experiments, landmark markers are positioned within a 4m by 4m flight area (see Fig. \ref{fig:test_venue}). For the experiment, the UAV flew at a constant height of \textit{0.8m} and the maximum speed is limited to \textit{0.3m/s}. 

\begin{table}[htb]
    \vspace{0.5cm}
    \setlength{\extrarowheight}{0.9ex}
    \caption{\small \textbf{Comparison of localisation performance across numerous trajectories:} We report the mean square error $({m}^2)$ and its standard deviation  from the VICON position (lower is better).}
    \vspace{-0.5cm}
    \begin{center}
    \vspace{0.2cm}
    \small
    \begin{tabularx}{\linewidth}{>{\centering\arraybackslash}X>{\centering\arraybackslash}X>{\centering\arraybackslash}X>{\centering\arraybackslash}X}
        \hline
        \toprule
        \hline
        \small \textbf{Landmark configuration} & \small \textbf{Box} & \small \textbf{Circle} & \small \textbf{Figure 8} \\
        \hline
        \small No Tag & \small 0.25($\pm$0.23) & \small 0.24($\pm$0.25) & \small 0.64($\pm$0.43) \\
        \small 1 tag & \small 0.19($\pm$0.17) & \small 0.20($\pm$0.19) & \small 0.26($\pm$0.26) \\
        \small 2 tags & \small \textbf{0.16($\pm$0.18)} & \small \textbf{0.13($\pm$0.17)} & \small \textbf{0.19($\pm$0.18)} \\
        \hline
    \end{tabularx}
    \end{center}
    \vspace{-0.5cm}
    \label{table:1}
\end{table}

\subsection{Position estimate performance}
We assess the position estimate error by comparing the VICON estimates with those from the landmark localization module. The mean square error of these comparisons is calculated and presented in Table \ref{table:1}. Additionally, we conducted ablation tests to evaluate the impact of the number of markers and its configurations on position estimate accuracy. Specifically, we tested three configurations: no markers, a single marker, and dual markers (each the size of an A4 sheet) at each landmark point. The outcomes of these tests are also summarized in Table \ref{table:1}. 

Our ablation study results demonstrate that increasing the number of landmarks reduces the position estimate error of the box and circle trajectories by \textbf{36.0\%} and \textbf{45.8\%} respectively. Notably, for the Figure 8 trajectory with 2 tags configuration, achieve improvements of \textbf{70.3\%}. We observed that pure optical flow tends to diverge on more complex trajectories, whereas the landmark localization module maintains consistent position estimation performance, demonstrating its robustness in challenging flight paths.

However, it is important to note that the improvement in position error is not directly proportional to the number of markers per landmark. This non-linear relationship likely stems from a trade-off between the benefits and drawbacks of adding more markers. While additional markers provide more data points, enhancing the landmark localization module's ability to correct positions accurately, they also introduce more noise into the system. Given that marker detection relies on vision-based methods, which can be error-prone, each extra marker increases the likelihood of detection errors. Consequently, although more markers have the potential to improve accuracy, the associated noise can offset these gains, resulting in diminishing improvements.

\begin{figure}
    \centering
    \includegraphics[width=0.4\textwidth]{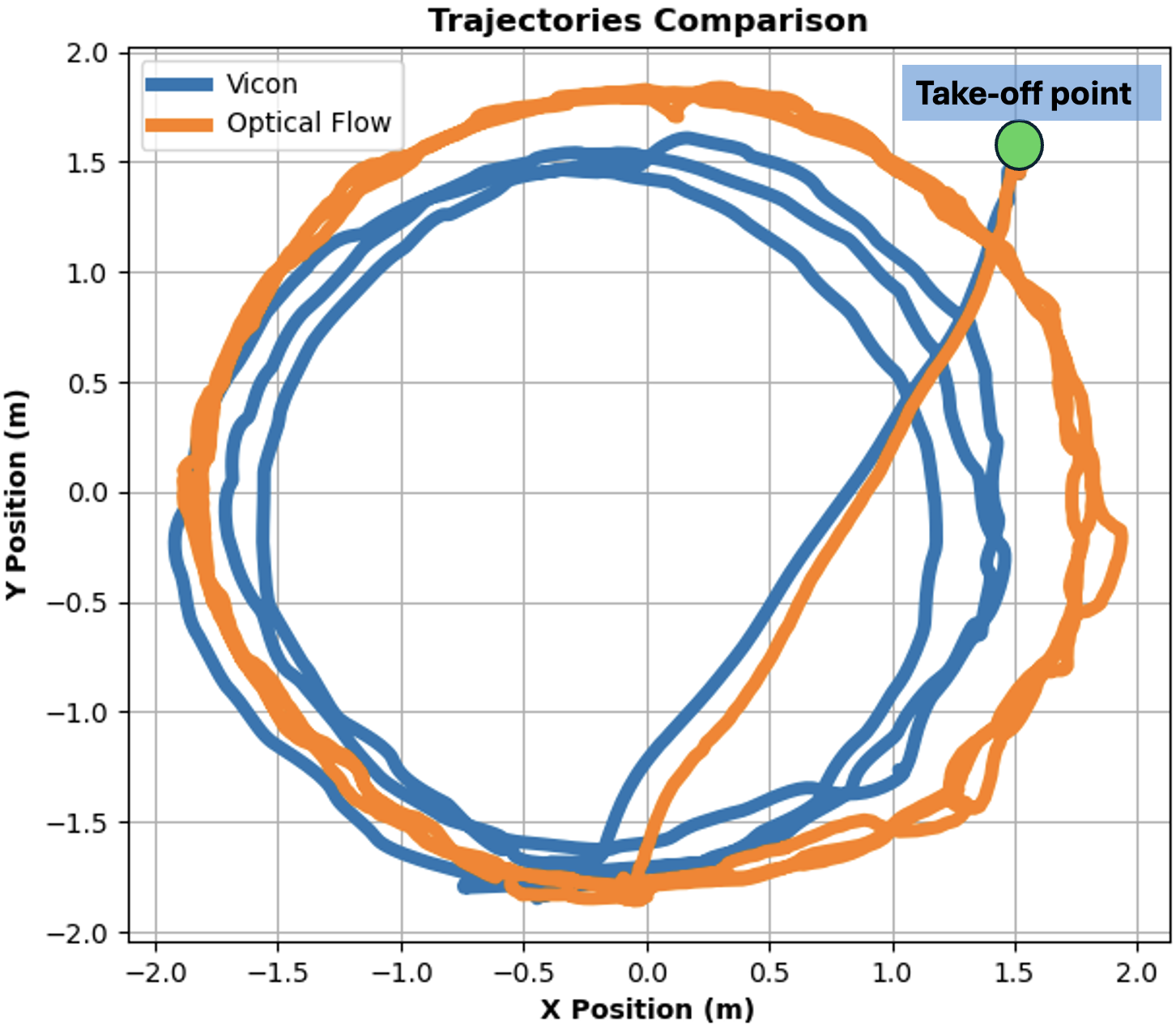}
    \caption{
    \textbf{Trajectory comparison for circle trajectory with 2 tags.} Ground truth position is represented by a blue solid line, while the landmark-localized position is shown by an orange solid line.}
    \label{fig: results}
    \vspace{-0.5cm}
\end{figure}

It is important to highlight that the pose estimate adjusts itself upon detecting fiducial markers, as illustrated in Fig. \ref{fig: results}. The irregularities in the paths suggest that position corrections is occurring during the experiments. This validation highlights the effectiveness of the landmark location module in helping the UAV correct its position estimates and limit localization errors. Without this module, localization errors could accumulate significantly, resulting in a drift that makes the localization unreasonable for normal flight, as demonstrated in the Figure 8 test trajectory experiment. Our landmark localization approach ensures stable localization performance in physical aerial swarm experiments. Additionally, the system allows for the occlusion of some landmark points, meaning that not all landmarks are necessary for position correction. This flexibility allows researchers to design experiments with cluttered obstacles, which is challenging with a motion capture system. 

 \subsection{System latency}

The overall system performance is affected by the latency in receiving corrected position estimates. To analyze this, we collected data presented in Table \ref{table:2}. The image capture rate measures how quickly the camera sensor captures an image, while the image transfer rate shows how often the image is sent to the ground computer for processing. The marker processing time, which is 0.163 seconds (or 6Hz), indicates the duration to compute a corrected position. Given the image capture rate of 15Hz, the bottleneck is the marker processing time, leading to a landmark-based localization correction frequency of 6Hz. This is a reasonable frequency for position corrections.

 \begin{table}[h]
    \vspace{0.1cm}
    \setlength{\extrarowheight}{0.9ex}
    \caption{\small \textbf{System latency metrics}}
    \vspace{-0.3cm}
    \begin{center}
    \begin{tabu}to\linewidth{X[c]X[c]X[c]}\hline
    \toprule
    \hline
     \small \textbf{Image capture time (ms)} & \small \textbf{Image transfer rate (Hz) } & \small \textbf{Marker processing time (s)}\\
    \hline
    \small 65-66 & \small 8-9 & \small 0.163 \\ 
    \hline
    \end{tabu}
    \end{center}
    \vspace{-0.7cm}
    \label{table:2}
\end{table}


\section{Conclusion}
\label{sec:conclusion}
This paper introduces a novel framework to the research community, providing support for advances in aerial swarm robotics tasks. The framework's key attributes include versatile and modular setup options that are affordable, while being independent of expensive motion capture system. WWe designed STAR to cover the entire mission spectrum, starting from simultaneous take-off and ending with landing, facilitated by a unified task manager. We also achieved stable and reasonable localization through our landmark localization approach. We validated STAR through  real-world experiments, validating its position estimates as well as various system performance. In the future, our goal is to develop a swarm framework that allows physical UAVs to join or leave the team dynamically. The ability of UAVs to join or exit the swarm is crucial in scenarios where they may be needed to enhance the capabilities of the team to complete their mission or undergo maintenance. Additionally, we plan to explore peer-to-peer communication among UAVs to establish a truly decentralized swarm framework, thus enhancing research efforts in decentralized aerial swarm tasks. Our ongoing objective is to further enhance the STAR framework, creating a robust platform poised to benefit the aerial swarm research community.

\bibliographystyle{IEEEtran}
\bibliography{citations}


\end{document}